%% file: main.tex
\documentclass[letterpaper]{article} 
\usepackage{aaai2026}  
\usepackage{times}  
\usepackage{helvet}  
\usepackage{courier}  
\usepackage[hyphens]{url}  
\usepackage{graphicx} 
\usepackage{natbib}  
\usepackage{caption} 
\usepackage{algorithm}
\usepackage{algorithmic}
\usepackage{booktabs}
\usepackage{amsmath} 
\usepackage{multirow}

\usepackage{newfloat}
\usepackage{listings}
\DeclareCaptionStyle{ruled}{labelfont=normalfont,labelsep=colon,strut=off} 
\floatstyle{ruled}
\newfloat{listing}{tb}{lst}{}
\floatname{listing}{Listing}
\definecolor{mypink}{RGB}{239,43,159}

\title{TraveLLaMA: A Multimodal Travel Assistant with \\Large-Scale Dataset and Structured Reasoning\\
\vspace{1.5mm}
{\normalsize\color{mypink}
Project: \url{https://travellama-best.github.io/}}
}
\author{Meng Chu\textsuperscript{\rm 1}, Yukang Chen\textsuperscript{\rm 2}, Haokun Gui\textsuperscript{\rm 1}, Shaozuo Yu\textsuperscript{\rm 2}, Yi Wang\textsuperscript{\rm 3}, Jiaya Jia\textsuperscript{\rm 1}}
\affiliations{
    \textsuperscript{\rm 1}Hong Kong University of Science and Technology\\

    \textsuperscript{\rm 2}Chinese University of Hong Kong\\
    \textsuperscript{\rm 3}Shanghai AI Laboratory
}

\begin{document}

\maketitle

\input{sec/0_abstract}

\input{sec/1_intro}
\input{sec/2_related_work}
\input{sec/3_dataset}
\input{sec/4_method}

\input{sec/5_experiment}

\input{sec/6_conclusion}

\input{RC}
\bibliography{aaai2026}

\end{document}

%% file: sec/0_abstract.tex
\begin{abstract}
Tourism and travel planning increasingly rely on digital assistance, yet existing multimodal AI systems often lack specialized knowledge and contextual understanding of urban environments. We present \textbf{TraveLLaMA}, a specialized multimodal language model designed for comprehensive travel assistance. Our work addresses the fundamental challenge of developing practical AI travel assistants through three key contributions: (1) TravelQA, a novel dataset of 265k question-answer pairs combining 160k text QA from authentic travel sources, 100k vision-language QA featuring maps and location imagery, and 5k expert-annotated Chain-of-Thought reasoning examples; (2) Travel-CoT, a structured reasoning framework that decomposes travel queries into spatial, temporal, and practical dimensions, improving answer accuracy by 10.8\% while providing interpretable decision paths; and (3) an interactive agent system validated through extensive user studies. Through fine-tuning experiments on state-of-the-art vision-language models (LLaVA, Qwen-VL, Shikra), we achieve 6.2-9.4\% base improvements, further enhanced by Travel-CoT reasoning. Our model demonstrates superior capabilities in contextual travel recommendations, map interpretation, and scene understanding while providing practical information such as operating hours and cultural insights. User studies with 500 participants show TraveLLaMA achieves a System Usability Scale score of 82.5, significantly outperforming general-purpose models and establishing new standards for multimodal travel assistance systems.
\end{abstract}

%% file: sec/1_intro.tex
\section{Introduction}
\label{sec:intro}

Travel planning exemplifies the complexity of real-world AI applications, requiring simultaneous understanding of visual scenes, geographical contexts, and practical constraints. While large language models (LLMs) have achieved remarkable success in many domains \cite{touvron2023llama, zhao2023survey}, they struggle with travel assistance due to a critical gap: the absence of multimodal datasets that capture the inherently visual and contextual nature of travel planning \cite{schumann2024velma, li2024georeasoner}.
\begin{figure}[t!]
\centering

\includegraphics[width=0.99\linewidth]{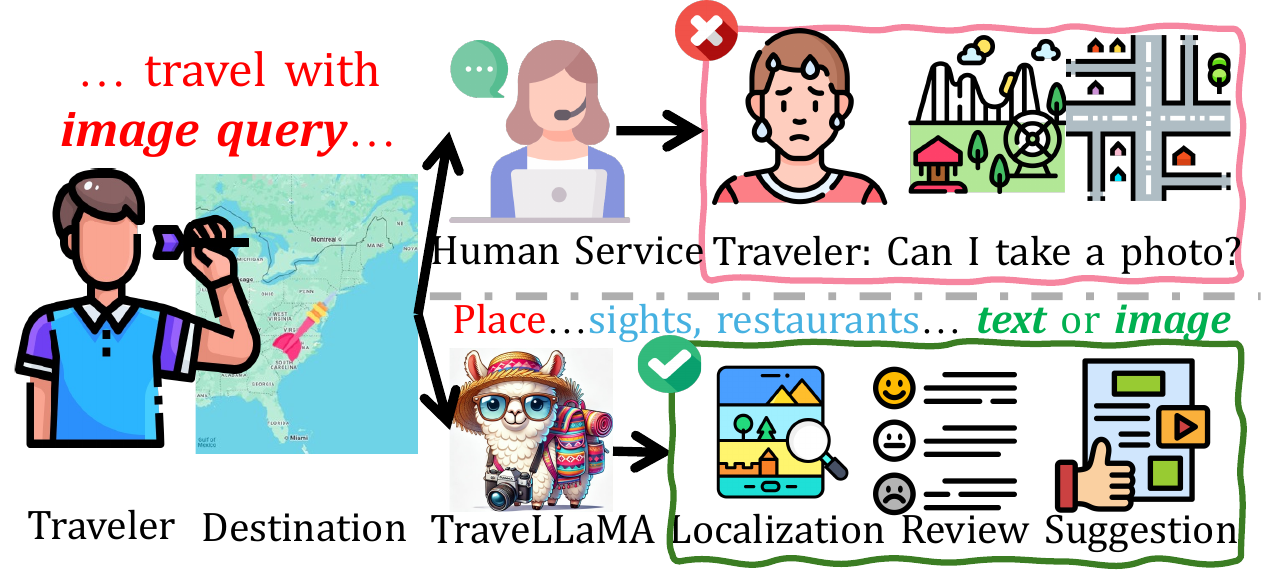} \\

\caption{TraveLLaMA, an advanced multimodal AI travel assistant that seamlessly processes both text and image-based queries. This powerful system enables travelers to plan trips efficiently by providing contextual responses including human service information, localization details, and personalized recommendations based on visual inputs and textual questions about destinations, sights, and restaurants.}
\label{fig:architecture}
\end{figure}

This limitation is particularly acute because effective travel assistance demands integration across multiple modalities. Recommending a restaurant, for instance, requires understanding its location on a map, recognizing ambiance from photos, interpreting user reviews, and considering operational constraints—a level of multimodal reasoning that current datasets fail to support \cite{yang2024v, vivanco2023geoclip}. Similarly, planning a day-long itinerary involves analyzing distances between attractions, understanding transportation options from visual maps, recognizing architectural styles from images, and incorporating temporal constraints like opening hours and peak times. General-purpose models, despite their broad training, lack the domain-specific knowledge to connect visual landmarks with practical travel information \cite{li2023geolm, maheshwary2024pretraining}.

Furthermore, the challenge extends beyond simple multimodal understanding to requiring culturally-aware and contextually-appropriate reasoning. A temple visit in Kyoto demands different preparations than a beach day in Bali, yet current AI systems often provide generic advice that fails to capture these nuances. The absence of structured reasoning frameworks means that even when models can process individual modalities, they struggle to synthesize information coherently—leading to recommendations that may be factually correct but practically infeasible, such as suggesting attractions without considering accessibility for elderly travelers or recommending outdoor activities without accounting for seasonal weather patterns.

\begin{figure*}[t!]
\centering

\includegraphics[width=0.98\linewidth]{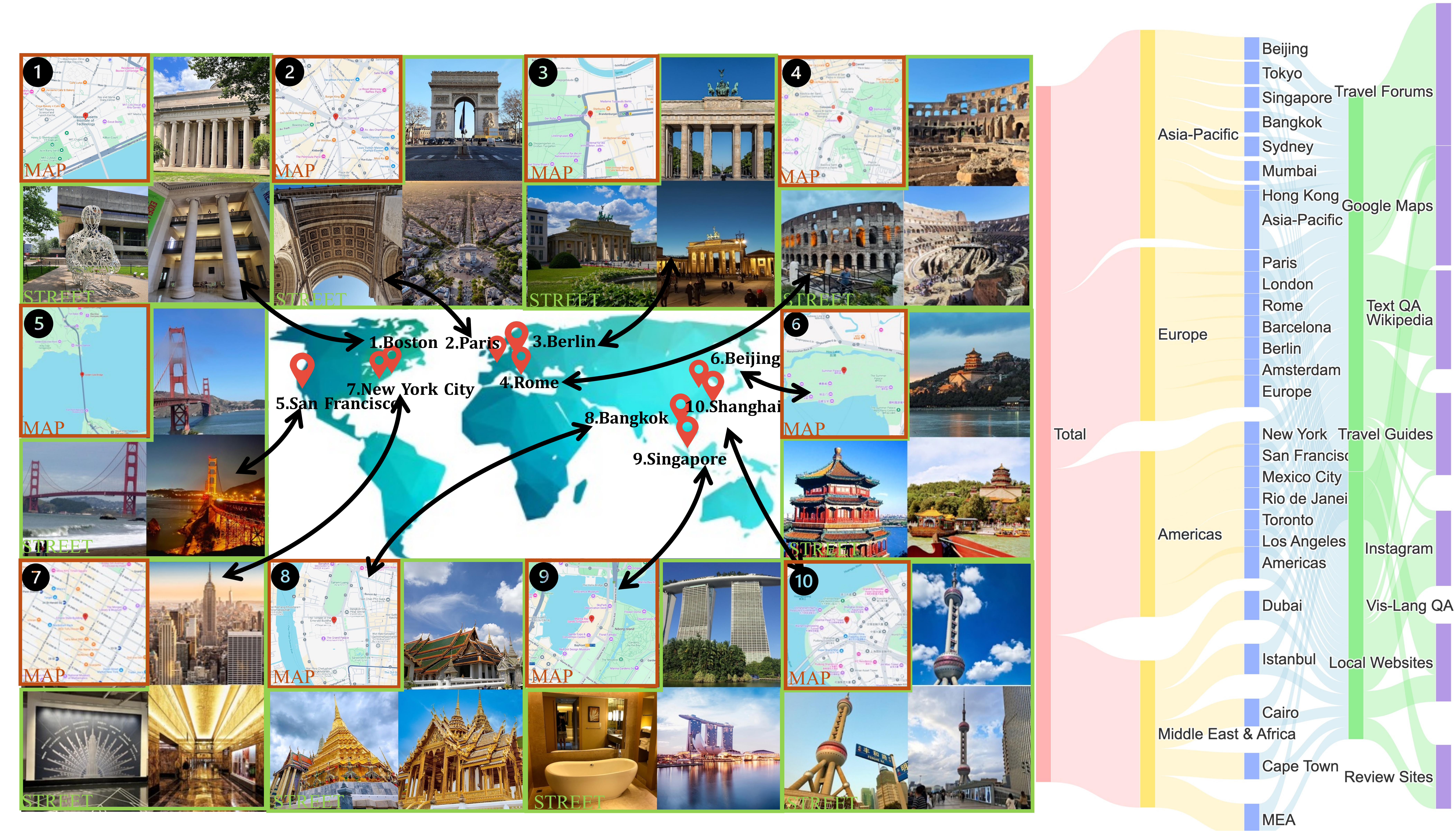} \\

\caption{The TravelQA dataset features iconic landmarks and destinations across major cities worldwide, connecting locations like San Francisco, New York, Paris, Rome, Berlin, Bangkok, Singapore, Shanghai, and Beijing through a global travel network.}
\label{fig:map}
\end{figure*}

While structured travel datasets remain scarce, the web contains abundant unstructured content across forums, review sites, and mapping services. We leverage modern LLMs to transform this fragmented information into high-quality multimodal training data \cite{xie2023quert}, enabling cost-effective dataset creation at scale.

To address these challenges, we present TraveLLaMA (Figure \ref{fig:architecture}), a specialized multimodal system for AI-powered travel assistance. Our contributions are:

\begin{itemize}
\item \textbf{TravelQA Dataset.} We create the first large-scale multimodal travel dataset with 265k QA pairs: 160k text-based pairs from travel forums, 100k vision-language pairs with maps and photos, and 5k expert-annotated CoT reasoning examples.

\item \textbf{Travel-CoT Reasoning.} Beyond base improvements of 6.2-9.4\%, our Travel-CoT framework decomposes queries into spatial, temporal, and practical dimensions, achieving 10.8\% accuracy gain with interpretable reasoning paths.

\item \textbf{Interactive Agent System.} Our ReAct-based agent integrates real-time services for dynamic planning, validated by 500 users with a SUS score of 82.5, demonstrating superior usability for complex travel tasks.
\end{itemize}

\begin{figure*}[t]
\centering

\includegraphics[width=0.99\linewidth]{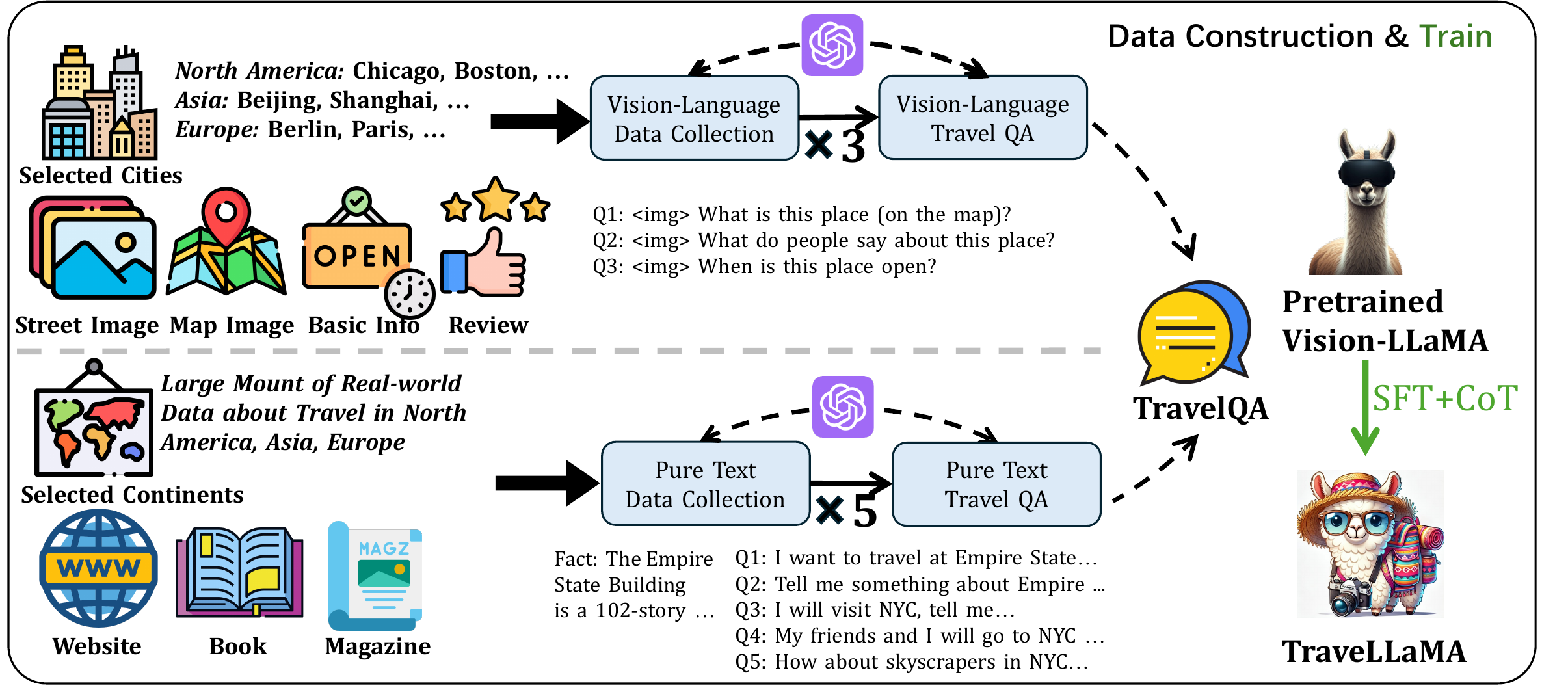} \\

\caption{TraveLLaMA's data construction and training process combines vision-language and text-based travel QA from diverse global sources through multi-round collection and fine-tuning.}
\label{fig:flow}
\end{figure*}

%% file: sec/2_related_work.tex
\section{Related Work}
\label{sec:related_work}

\noindent\textbf{Large Language Models.}
Large Language Models (LLMs) \cite{shafik2024introduction, touvron2023llama, zeng2022glm, zhao2023survey, chu2025graphvideoagent} show impressive capabilities yet underperform in specialized domains, prompting development of domain-specific models for finance \cite{wu2023bloomberggpt}, medicine \cite{singhal2023large}, mathematics \cite{azerbayev2023llemma}, and geospatial applications \cite{deng2024k2, wang2023optimizing}. These models have been studied for spatial understanding \cite{gurnee2023language, momennejad2023evaluating, yamada2023evaluating} and geographic representation \cite{godey2024scaling, manvi2023geollm}. LLM-powered language agents like AutoGPT \cite{yang2023auto}, BabyAGI \cite{nakajima2023task}, and HuggingGPT \cite{shen2023hugginggpt} decompose complex tasks through Memory, Tool-use, and Planning modules, utilizing memory summarization \cite{chen2023walking, zhou2023recurrentgpt, liang2023unleashing, Yu_2025_ICCV}, retrieval techniques \cite{andreas2022language, park2023generative, zhong2024memorybank}, and tool-augmentation \cite{nakano2021webgpt, lu2023chameleon, ge2023openagi, xie2023openagents}. Our approach focuses on understanding physical urban spaces and spatial reasoning to solve real urban challenges, demonstrating the geospatial knowledge capabilities in pre-trained language models.

\noindent\textbf{Vision-Language Models.} 
The evolution of vision-language models has seen significant advancement, beginning with early works in visual-semantic embedding \cite{frome2013devise, kiros2014unifying, chu2024towards} and progressing to more sophisticated architectures. Recent models like GPT-4V \cite{openai2023gpt4v} and LLaVA \cite{liu2023llava} have demonstrated unprecedented capabilities in joint text-vision understanding.
However, these general-purpose models face significant challenges in travel-specific applications. Recent work has attempted to address these issues through specialized architectures \cite{yang2023specialized} and domain-specific pre-training \cite{zhao2023domain}, but significant challenges remain in achieving human-level understanding of travel-related visual content \cite{brown2023limitations}.

\noindent\textbf{LLMs in Urban Applications.}
Travel AI systems have evolved from basic route planning and POI recommendation to advanced systems like GeoLLM \cite{li2023geolm} and GeoReasoner \cite{li2024georeasoner}. Recent work focuses on multi-modal integration \cite{yang2024v, vivanco2023geoclip}, though challenges persist in temporal reasoning and cross-modal alignment \cite{schumann2024velma}. Specialized developments span cultural context, accessibility, and personalization \cite{maheshwary2024pretraining}, while real-world implementations reveal challenges in uncertainty handling and system reliability. Applications extend to traffic safety analysis, geospace understanding, and accident reporting \cite{chu2024towards, feng2024citygpt}. While TravelPlanner \cite{xie2024travelplanner} focuses on text-only travel planning, our system addresses multimodal single-day urban itinerary planning with multi-day extensibility, motivated by current limitations in handling complex travel planning scenarios \cite{xie2023quert}.

\begin{table}[t]
\centering
\label{tab:dataset_composition}
\small
\begin{tabular}{llrrr}
\toprule
\textbf{Category} & \textbf{Subcategory} & \textbf{All} & \textbf{Train} & \textbf{Test} \\
\midrule
\multirow{4}{*}{QA Format} & Text & 160k & 128k & 32k \\
& Vis-Lang & 100k & 80k & 20k \\
& CoT & 5k & 5k & --\\
& \textbf{Total} & \textbf{265k} & \textbf{213k} & \textbf{52k}$^*$ \\
\midrule
\multirow{6}{*}{Locations} & Attractions & 70k & 56k & 14k \\
& Dining & 52k & 42k & 10k \\
& Living & 39k & 31k & 8k \\
& Transportation & 26k & 21k & 5k \\
& Cultural & 39k & 31k & 8k \\
& Practical & 34k & 27k & 7k \\
\midrule
\multirow{2}{*}{Visual Elements} & Map & 40k & 32k & 8k \\
& Street & 60k & 48k & 12k \\
\midrule
\multirow{3}{*}{CoT Annotations} & Spatial & 5k & 5k & -- \\
& Temporal & 5k & 5k & -- \\
& Practical & 5k & 5k & -- \\
\bottomrule
\end{tabular}
\caption{Complete Dataset Composition and Usage Analysis}
\end{table}

%% file: sec/3_dataset.tex
\begin{figure*}[t!]
\centering
\includegraphics[width=0.99\linewidth]{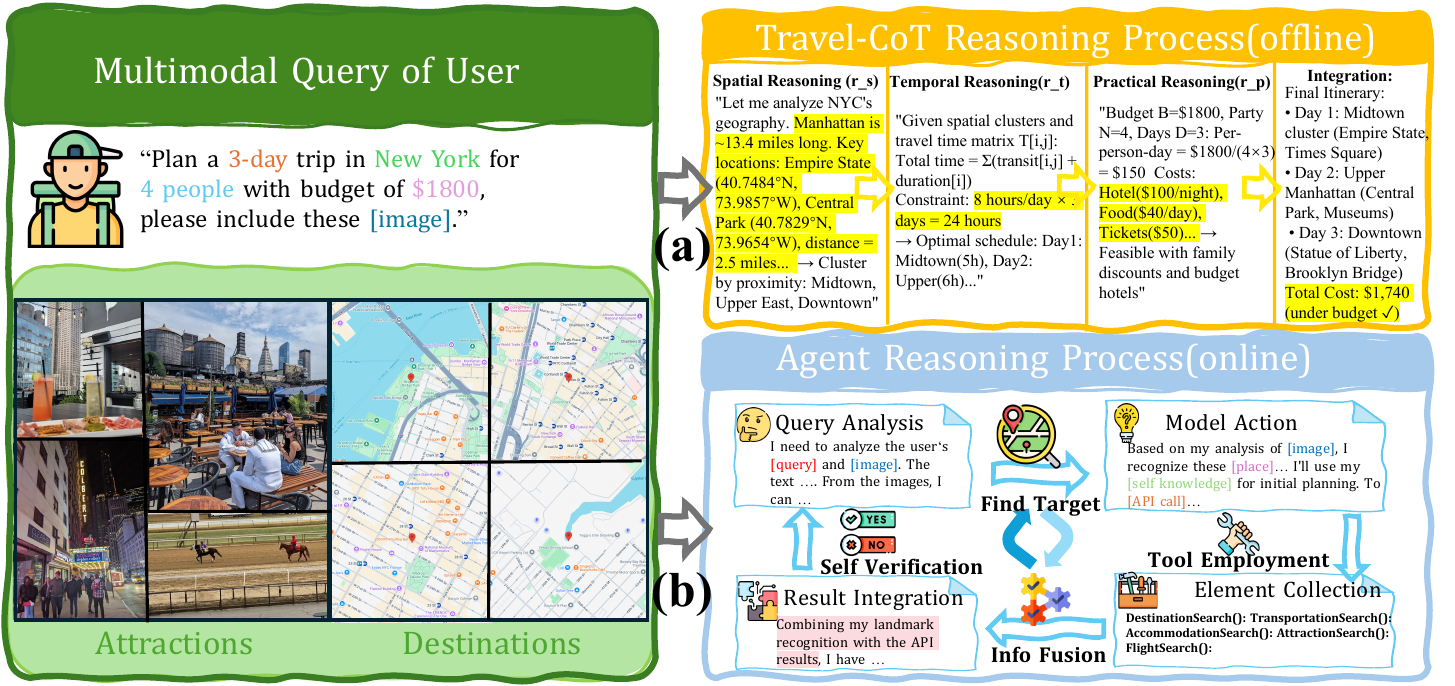}
\caption{TraveLLaMA uses a reasoning and acting process to create travel plans. When a user submits a text-image query, the system could do the reasoning process offline or online, analyzing both components, identifies locations, employs specialized tools through API calls, and generates detailed itineraries with budget calculations matching user requirements.}
\label{fig:agent}
\end{figure*}

\section{TravelQA}
\label{sec:related_work}

\subsection{Dataset Overview}

Figure~\ref{fig:map} and Table~1 present TravelQA, a multimodal resource containing 265k QA pairs. To clarify points raised by reviewers, we explicitly describe the numerical composition: the dataset includes 160k text-only QA pairs, 100k vision–language QA pairs, and 5k expert-annotated CoT examples. The 160k text QA pairs come from 26k factual units expanded into five diverse questions each (130k), plus 30k augmented QAs focusing on practical constraints such as safety, cost, and accessibility. The 100k vision–language QA pairs are constructed from 20k POIs, each with an average of 4–5 street-view or map images; GPT-4 generates three question types \emph{per image}, yielding roughly \(20\text{k} \times 4\text{–}5 \times 3 \approx 100\text{k}\) pairs. The 5k CoT examples represent a single set where each item contains spatial, temporal, and practical reasoning, rather than 15k separate labels. To prevent leakage across modalities, \textbf{all splits are POI-disjoint}: every POI and its associated images and metadata appear in exactly one split. The dataset spans 35+ cities across North America, Asia, and Europe, covering varied cultural and geographic contexts. Text-based answers average 45.6 words, while visual answers average 25–28 words, with all QA pairs passing multi-stage verification to ensure quality and factual consistency.

\subsection{Dataset Construction Methodology}

\noindent\textbf{Pipeline Overview.}
As shown in Figure~\ref{fig:flow}, TravelQA is built through parallel text and vision–language tracks processing street images, maps, descriptions, reviews, and structured facts. These supervised signals enable TraveLLaMA to answer a wide variety of travel queries.

\noindent\textbf{Vision–Language QA Construction.}
For each of the 20k POIs, we collect 4–5 images, descriptive metadata, operational information, and filtered reviews. GPT-4 generates three QA types per image—identification, experience, and practical—resulting in approximately 100k QA pairs. Answers average 25–28 words and capture both visual content and contextual information.

\noindent\textbf{Text-Based QA Generation.}
From 26k facts extracted from travel guides and online resources, GPT-4 produces five diverse questions per fact, alongside 30k additional QAs emphasizing practical constraints. Each QA pair undergoes a three-layer verification procedure combining rule checks, semantic consistency checks, and manual spot reviews, yielding 160k high-quality text QAs.

\noindent\textbf{Travel-CoT Data Construction.}
To teach structured reasoning, we annotate 5,000 complex queries requiring multi-factor decision making. Travel experts provide explicit chains-of-thought containing \emph{spatial}, \emph{temporal}, and \emph{practical} reasoning steps within each example, followed by final integrated answers. This unified annotation format ensures coherent multi-dimensional reasoning rather than separate labels.

\noindent\textbf{Discussion.}
Table~\ref{tab:dataset_comparison} highlights the breadth of TravelQA compared with prior resources. By integrating text, maps, street images, and structured CoT annotations, the dataset enables models to jointly reason over spatial layout, schedules, cultural context, and practical constraints—capabilities essential for real-world travel assistance systems.

\begin{table}[t]
\centering

\small
\begin{tabular}{lll}
\toprule
\textbf{Feature} & \textbf{TravelQA (Ours)} & \textbf{TravelPlanner} \\
\midrule
Modality & Text+Visual & Text-only \\
Query Scale & 265k & 1,225 \\
Visual Content & Map, Street Photos & None \\
Category & 6 structured categories & Unstructured \\
& (Attractions, Dining, etc.) & \\
\bottomrule
\end{tabular}
\caption{Comparative Analysis of Datasets}\label{tab:dataset_comparison}
\end{table}

%% file: sec/4_method.tex
\begin{table*}[t!]
\fontsize{12pt}{10pt}\selectfont
\centering
\resizebox{\linewidth}{!}{
\begin{tabular}{l l c  c | c c | c c | c c}
\toprule
\multirow{2}{*}{Method} & \multirow{2}{*}{LLM} & Image & \multicolumn{1}{c|}{Sample} & \multicolumn{2}{c|}{Pure Text} & \multicolumn{2}{c|}{VQA} & \multicolumn{2}{c}{Full} \\
 & & Size & Size & Score & $\Delta$ & Score & $\Delta$ & Score & $\Delta$ \\
\midrule
\multicolumn{10}{l}{\textit{Pretrained Models}} \\
BLIP-2~\cite{blip-2} & Vicuna-13B~\cite{vicuna2023} & 224$^2$ & 129M  & 60.3 & - & 51.6 & - & 56.9 & - \\
InstructBLIP~\cite{instructblip} & Vicuna-7B~\cite{vicuna2023} & 224$^2$ & 129M  & 62.8 & - & 54.1 & - & 59.4 & - \\
InstructBLIP~\cite{instructblip} & Vicuna-13B~\cite{vicuna2023} & 224$^2$ & 129M  & 64.6 & - & 55.4 & - & 61.1 & - \\
Shikra~\cite{shikra} & Vicuna-13B~\cite{vicuna2023} & 224$^2$ & 600K  & 71.6 & - & 60.8 & - & 67.5 & - \\
Qwen-VL~\cite{qwen-vl} & Qwen-7B~\cite{bai2023qwen} & 448$^2$ & 1.4B & 72.1 & - & 61.6 & - & 68.1 & - \\
Qwen-VL-Chat~\cite{qwen-vl} & Qwen-7B~\cite{bai2023qwen} & 448$^2$ & 1.4B & 73.2 & - & 62.8 & - & 69.2 & - \\
LLaVA-1.5~\cite{llava} & Vicuna-7B~\cite{vicuna2023} & 336$^2$ & 558K & 72.8 & - & 62.3 & - & 68.8 & - \\
LLaVA-1.5~\cite{llava}& Vicuna-13B~\cite{vicuna2023} & 336$^2$ & 558K  & 74.3 & - & 63.3 & - & 70.0 & - \\
\midrule
\multicolumn{10}{l}{\textit{Fine-tuned Models}} \\
BLIP-2~\cite{blip-2} & Vicuna-13B~\cite{vicuna2023} & 224$^2$ & 129M  & 64.7 & \textbf{+7.3\%} & 54.7 & \textbf{+6.0\%} & 60.9 & \textbf{+7.0\%} \\
InstructBLIP~\cite{instructblip} & Vicuna-7B~\cite{vicuna2023} & 224$^2$ & 129M & 68.2 & \textbf{+8.6\%} & 58.2 & \textbf{+7.6\%} & 64.4 & \textbf{+8.4\%} \\
InstructBLIP~\cite{instructblip} & Vicuna-13B~\cite{vicuna2023} & 224$^2$ & 129M  & 68.8 & \textbf{+6.5\%} & 58.8 & \textbf{+6.1\%} & 64.9 & \textbf{+6.2\%} \\
Shikra~\cite{shikra} & Vicuna-13B~\cite{vicuna2023} & 224$^2$ & 600K & 77.7 & \textbf{+8.5\%} & 66.7 & \textbf{+9.7\%} & 73.5 & \textbf{+8.9\%} \\
Qwen-VL~\cite{qwen-vl} & Qwen-7B~\cite{bai2023qwen} & 448$^2$ & 1.4B & 78.7 & \textbf{+9.2\%} & 67.7 & \textbf{+9.9\%} & 74.5 & \textbf{+9.4\%} \\
Qwen-VL-Chat~\cite{qwen-vl} & Qwen-7B~\cite{bai2023qwen} & 448$^2$ & 1.4B & 78.4 & \textbf{+7.1\%} & 67.4 & \textbf{+7.3\%} & 74.2 & \textbf{+7.2\%} \\
LLaVA-1.5~\cite{llava} & Vicuna-7B~\cite{vicuna2023} & 336$^2$ & 558K & 78.0 & \textbf{+7.1\%} & 67.0 & \textbf{+7.5\%} & 73.8 & \textbf{+7.3\%} \\
LLaVA-1.5~\cite{llava}& Vicuna-13B~\cite{vicuna2023} & 336$^2$ & 558K  & 80.4 & \textbf{+8.2\%} & 68.9 & \textbf{+8.8\%} & 76.0 & \textbf{+8.6\%} \\
\midrule
\multicolumn{10}{l}{\textit{Fine-tuned Models with Travel-CoT}} \\
TraveLLaMA (Ours) & Vicuna-13B~\cite{vicuna2023} & 336$^2$ & 558K  & 82.5 & \textbf{+10.7\%} & 70.5 & \textbf{+11.0\%} & 77.8 & \textbf{+10.8\%} \\
\bottomrule
\end{tabular}
}
\caption{Comparisons of performance scores (Pure Text, VQA, and Full scores) and improvements across different vision-language models, showing both pretrained and fine-tuned versions with their respective LLM backbones, image sizes, and sample sizes. Full score is computed as weighted average based on test set distribution (61.5\% text, 38.5\% vision-language).}
\label{tab:results}
\end{table*}

\section{Method}

We develop TraveLLaMA by augmenting vision–language models with structured reasoning (Travel-CoT) and an agentic architecture tailored for real-world travel assistance.

\subsection{Travel-CoT: Structured Reasoning Enhancement}

Pre-trained VLMs answer factual queries well but struggle with planning queries that require multi-factor reasoning. To address this, Travel-CoT explicitly decomposes each query into spatial, temporal, and practical reasoning steps. Following reviewer feedback, we adopt a two-stage formulation that better reflects our true implementation. Given multimodal input $(x, Q)$, the model first generates a reasoning chain
\[
r = f_{\theta}(x, Q), \quad r = \{r_s, r_t, r_p\},
\]
where $r_s$ encodes spatial understanding (locations, distances, routes), $r_t$ encodes temporal scheduling (operating hours, time allocation), and $r_p$ captures practical constraints (budget, accessibility, safety). The final answer is then generated conditioned on both the input and the reasoning chain:
\[
y \sim P_{\phi}(y \mid x, Q, r).
\]
We train the two components jointly using supervised learning on 5,000 expert-annotated Travel-CoT examples:
\[
\mathcal{L}
= \lambda\, \mathcal{L}_{\mathrm{CoT}}(r^{\text{*}}, r)
+ \mathcal{L}_{\mathrm{ans}}(y^{\text{*}}, y).
\]

This structured reasoning improves plan correctness and interpretability while maintaining training efficiency through parameter-efficient tuning.

\subsection{Agent Architecture for Real-time Planning}

For practical deployment, we integrate Travel-CoT into a ReAct-style agent capable of processing multimodal travel requests with iterative reasoning and tool use (Figure~\ref{fig:agent}). The agent proceeds in four stages. First, \textit{Query Analysis} extracts textual constraints (destination, days, budget, group size) and interprets visual inputs such as uploaded photos for landmark recognition. Second, \textit{Reasoning} applies Travel-CoT to organize spatial, temporal, and practical requirements into an actionable planning state.

Next, the agent enters \textit{Tool Employment}, calling APIs for schedules, prices, reviews, and transit information. Its internal state evolves as:
\[
\text{Plan}_t = \text{Update}\big(\text{Plan}_{t-1},\, \text{Tool}(\pi(s_t, \text{Plan}_{t-1})),\, r\big),
\]
where $\pi$ selects appropriate tools based on the current reasoning state. Returned information (e.g., hours, prices, availability) is merged with visual grounding from the input images.

Finally, the agent executes \textit{Result Integration}, generating the final plan:
\[
y = \text{Generate}(\text{Plan}_T, \{o_1, \ldots, o_T\}, r),
\]
producing detailed itineraries with schedules, budgets, and constraint checks. The agent verifies compliance with user requirements (e.g., total cost, hours, accessibility) before returning the answer. For example, given an uploaded image of the Brooklyn Bridge and a budget limit, the agent identifies the landmark, retrieves hours and nearby attractions, and synthesizes a full-day schedule with cost calculations and timing alignment.

%% file: sec/5_experiment.tex
\begin{table}[t]
\small
\centering
\setlength{\tabcolsep}{4pt} 
\begin{tabular}{llcc}
\hline
\multicolumn{4}{l}{\textbf{A. Overall SUS Scores}} \\
\hline
& \textbf{System} & \textbf{SUS Score} & \textbf{Rating} \\
\cline{2-4}
& Ours & \textbf{82.5} & \textbf{Excellent} \\
& Claude 3.5 & 76.3 & Good \\
\hline
\multicolumn{4}{l}{\textbf{B. Items (0-10): Odd=(score-1)×2.5, Even=(5-score)×2.5}} \\
\hline
\textbf{Item} & \textbf{Description} & \textbf{Ours} & \textbf{Claude 3.5} \\
\hline
1 & Would use frequently $\uparrow$ & \textbf{8.8} & 8.0 \\
2 & Unnecessarily complex $\downarrow$ & \textbf{8.5} & 7.3 \\
3 & Easy to use $\uparrow$ & \textbf{8.6} & 7.8 \\
4 & Need technical support $\downarrow$& \textbf{8.8} & 7.5 \\
5 & Functions well integrated $\uparrow$ & \textbf{8.0} & 7.7 \\
9 & Confident using $\uparrow$& 7.9 & \textbf{8.0} \\
6 & Too much inconsistency $\downarrow$& \textbf{8.3} & 6.9 \\
10 & Needed to learn a lot $\downarrow$& 7.9 & \textbf{7.8} \\
7 & Quick to learn $\uparrow$& \textbf{8.7} & 8.1 \\
8 & Cumbersome to use $\downarrow$& \textbf{8.0} & 7.2 \\
\hline
\multicolumn{4}{l}{\textbf{C. Supplementary (1-5)}} \\
\hline
\textbf{Item} & \textbf{Question} & \textbf{Ours} & \textbf{Claude 3.5} \\
\hline
11 & Recommendations met needs & \textbf{4.5} & 4.1 \\
12 & Time Conservation & \textbf{4.6} & 4.2 \\
13 & Trust in recommendations & 4.3& \textbf{4.4} \\
14 & Would recommend to others & \textbf{4.5} & 4.0 \\
\hline
\multicolumn{4}{l}{\textbf{D. Participant Demographics}} \\
\hline
\multicolumn{2}{l}{Total participants} & \multicolumn{2}{c}{500 (250 per system)} \\
\multicolumn{2}{l}{Age range} & \multicolumn{2}{c}{18-62 years} \\
\hline
\end{tabular}
\caption{System Usability Scale (SUS) Evaluation Results.}\label{tab:sus}
\end{table}

\section{Experiments}
\label{sec:experiments}

\subsection{Experimental Setup}

\noindent\textbf{Dataset Configuration.} We divide TravelQA into training and testing sets with a 4:1 ratio, ensuring all splits are POI-disjoint across text, images, and metadata to prevent leakage. For evaluation, each test question is converted into a four-choice multiple-choice (MCQ) format to obtain precise, reproducible accuracy metrics. Distractors are constructed from semantically plausible alternatives within the same category or region (e.g., nearby POIs, similar attractions), rather than random sampling. Models generate free-form answers, which are mapped to MCQ options using normalized string matching and a lightweight semantic matcher; responses failing to match any option are counted as incorrect. Our assessment spans pure text understanding (travel queries, cultural knowledge), map comprehension for spatial layout and navigation cues, scene understanding for visual and architectural attributes, information extraction for operational details, review comprehension for synthesizing user-generated opinions, and temporal reasoning for operating hours and seasonal constraints. This comprehensive MCQ-based protocol provides a robust and standardized evaluation across all aspects of travel assistance.

\noindent\textbf{Training Details.} The training process follows a rigorous protocol designed to maximize model performance while ensuring reproducibility using 8 A100 GPUs. Our training dataset comprises 213k carefully curated QA pairs, representing a diverse range of travel-related scenarios and query types. The test set, consisting of 52k QA pairs, maintains a balanced distribution across different question categories and difficulty levels. We use normal QA for finetuning initally and then ues the CoT QA for post-training. All visual inputs are processed at a standardized resolution of 336 × 336 pixels, chosen to optimize the balance between computational efficiency and retention of important visual details. Text inputs are limited to a maximum length of 512 tokens, ensuring comprehensive coverage of complex travel queries while maintaining practical processing efficiency. The training process implements standard optimization techniques, including learning rate scheduling, gradient clipping, and early stopping based on validation performance.

\noindent\textbf{Evaluation Tasks.} Our evaluation combines quantitative QA accuracy, large-scale human usability assessment, and qualitative analysis. For quantitative testing, we evaluate TraveLLaMA on 52k POI-disjoint test questions in the TravelQA benchmark, covering both text-only queries (cultural information, travel advice) and vision–language tasks (landmark recognition, scene interpretation, and map-based spatial reasoning). Each test item is formatted as a four-choice MCQ, enabling precise measurement of model correctness. To assess real-world usability, we conduct a large-scale user study with 500 participants using the System Usability Scale (SUS). The study adopts a between-subjects design with anonymized system identities (blind evaluation). Participants complete four multi-step travel-planning tasks—including itinerary construction, constraint satisfaction, and iterative refinement—and then rate usability on the 10-item SUS questionnaire. We follow the standard SUS scoring procedure (1–5 Likert transformed to 0–100) and report results with 95\% confidence intervals, effect size (Cohen’s d), and statistical significance using a two-sided t-test. Finally, we perform qualitative comparisons between TraveLLaMA and strong baselines across representative travel scenarios, examining factual accuracy, constraint adherence, and practical utility.

\noindent\textbf{Performance Comparison.} Our experimental results in Table \ref{tab:results} demonstrate substantial improvements through domain-specific fine-tuning. Among pretrained models, LLaVA-1.5 with Vicuna-13B achieves the highest baseline performance (Pure Text: 74.3, VQA: 63.3, Full: 70.0). Fine-tuning on TravelQA yields consistent gains across all architectures, with improvements ranging from +6.2\% to +9.4\%. Notably, Qwen-VL shows the largest relative improvements (+9.2\% Pure Text, +9.9\% VQA, +9.4\% Full), while LLaVA-1.5 (Vicuna-13B) achieves the best absolute performance among fine-tuned baselines (80.4, 68.9, 76.0). Our complete TraveLLaMA system, incorporating Travel-CoT reasoning, further enhances performance to 82.5 Pure Text (+10.7\%), 70.5 VQA (+11.0\%), and 77.8 Full (+10.8\%) compared to the pretrained baseline. The results also reveal that models with larger image resolutions (448² for Qwen-VL vs 224² for BLIP-2) achieve better performance.

\noindent\textbf{Agent Performance Analysis.} Table~\ref{tab:sus} shows that TraveLLaMA achieves markedly higher usability than Claude3.5, with an SUS score of 82.5 (“Excellent”) versus 76.3 (“Good”). The 6.2-point gain is driven by TraveLLaMA’s domain-optimized design, reflected in strong ease-of-use, learnability, and reduced-complexity ratings. Users consistently found TraveLLaMA more intuitive and less cognitively demanding for travel planning. Claude3.5 shows small advantages in confidence and trust, likely influenced by general familiarity with the model. Based on a large-scale study of 500 users, the results indicate that a specialized travel agent with structured reasoning (Travel-CoT) provides a significantly better user experience than a general-purpose chatbot, particularly in minimizing cognitive load and improving task efficiency.

\begin{figure*}[t!]
\centering

\includegraphics[width=0.80\linewidth]{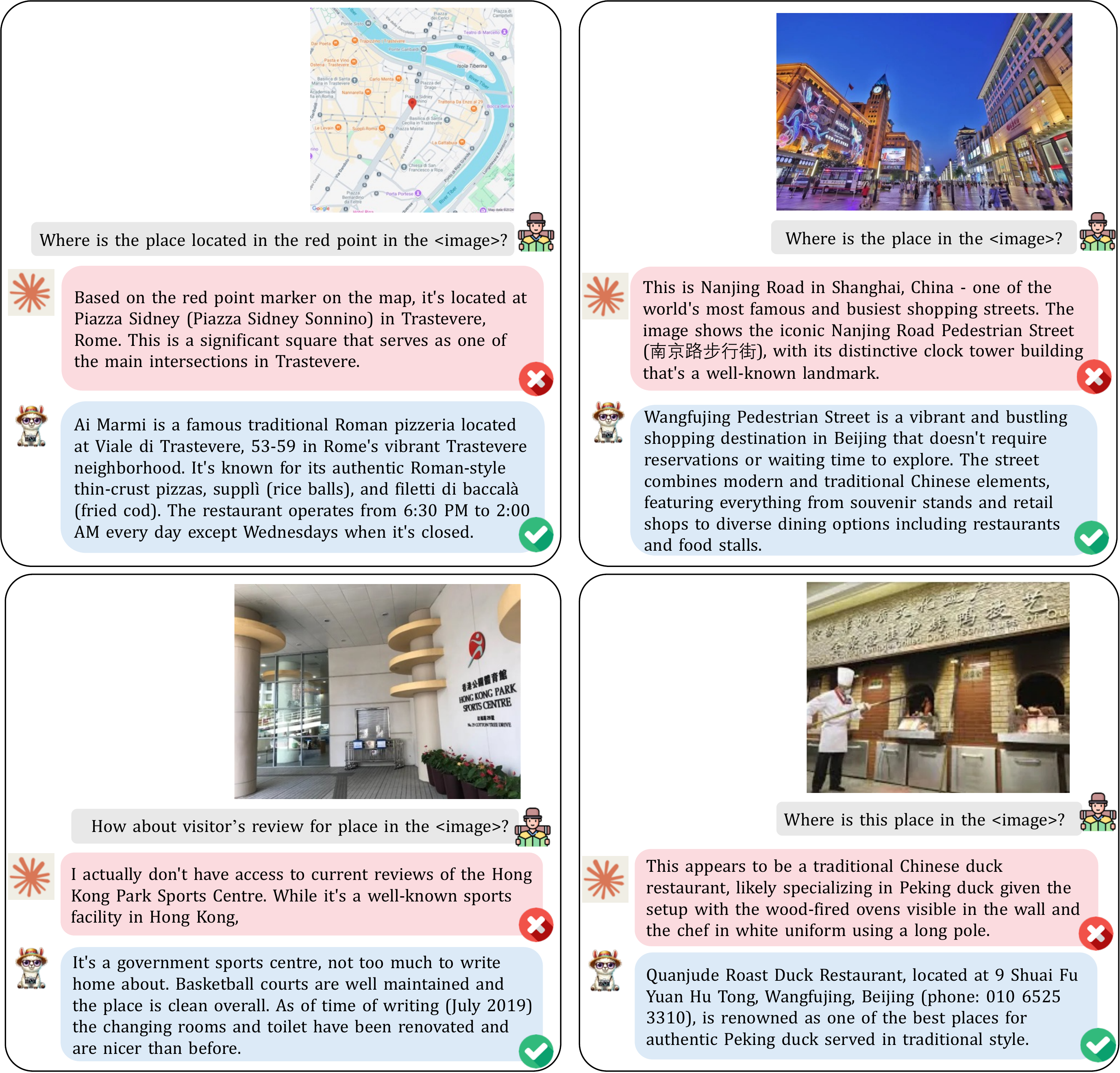} \\

\caption{Comparison between TraveLLaMA and Claude 3.5 shows that TraveLLaMA provides more accurate and detailed travel information in these location-based examples.}
\label{fig:comparetoclaude3.5}
\end{figure*}

\noindent\textbf{Qualitative Analysis.} Figure~\ref{fig:comparetoclaude3.5} illustrates that TraveLLaMA consistently outperforms Claude3.5 in accuracy and contextual understanding across diverse travel-related queries. In map-based tasks, TraveLLaMA shows precise location grounding—for example, correctly identifying Piazza Sidney in Rome’s Trastevere district and providing relevant nearby information such as local restaurants and operating hours—while Claude3.5 often fails to recognize the correct place. In scene recognition, TraveLLaMA accurately identifies iconic landmarks like Shanghai’s Nanjing Road and offers meaningful contextual descriptions, whereas Claude3.5 mislabels it as Wangfujing Street in Beijing. TraveLLaMA also demonstrates stronger multimodal integration, delivering detailed, actionable establishment information for venues such as the Hong Kong Park Sports Centre and Quanjude Restaurant, where Claude3.5 gives incomplete or incorrect details. The performance gap widens in more complex queries requiring both visual reasoning and domain knowledge: TraveLLaMA not only recognizes architectural styles and venue-specific constraints but also proactively provides helpful travel guidance, including nearby attractions or optimal visiting times. These observations highlight the advantages brought by domain-specific tuning and structured reasoning in TraveLLaMA.

%% file: sec/6_conclusion.tex
\section{Conclusion}

We presented TraveLLaMA, a specialized multimodal language model that advances AI-powered travel assistance through three key contributions: (1) TravelQA, the first large-scale dataset with 265k QA pairs combining text, images, and maps for travel-specific training; (2) Travel-CoT, a structured reasoning framework that decomposes queries into spatial, temporal, and practical dimensions, achieving 10.8\% accuracy improvement while providing interpretable decision paths; and (3) an interactive agent system that integrates real-time services for dynamic planning, validated through user studies with 500 participants achieving a SUS score of 82.5. Our comprehensive experiments demonstrate that domain-specific training yields 6.2-9.4\% improvements over general-purpose models, with Travel-CoT providing substantial additional gains. Beyond establishing effective methods for travel AI, this work offers valuable insights for developing specialized assistants in other complex multimodal domains. Future research directions include real-time information integration, multilingual support, and personalized planning mechanisms to further enhance the travel experience.

%% file: RC.tex
\section*{Acknowledgements}
This work was supported in part by the Research Grants Council under the Areas of Excellence scheme grant AoE/E-601/22-R.

%% file: aaai2026.bib
@String(ICCV= {Int. Conf. Comput. Vis.})

@String(AAAI = {AAAI})

@String(ICCV  = {ICCV})

@article{frome2013devise,
  title={DeViSE: A Deep Visual-Semantic Embedding Model},
  author={Frome, Andrea and others},
  journal={Advances in Neural Information Processing Systems},
  year={2013}
}

@article{kiros2014unifying,
  title={Unifying Visual-Semantic Embeddings},
  author={Kiros, Ryan and others},
  journal={arXiv preprint arXiv:1411.2539},
  year={2014}
}

@article{openai2023gpt4v,
  title={GPT-4V: Visual Capabilities in GPT-4},
  author={OpenAI},
  year={2023}
}

@article{liu2023llava,
  title={LLaVA: Large Language and Vision Aligned Model},
  author={Liu, Yang and others},
  journal={Proceedings of the IEEE/CVF Conference on Computer Vision and Pattern Recognition},
  year={2023}
}

@article{yang2023specialized,
  title={Specialized Architectures for Vision-Language Models in Tourism},
  author={Yang, Kai and others},
  journal={arXiv preprint arXiv:2306.10456},
  year={2023}
}

@article{zhao2023domain,
  title={Domain-Specific Pre-training for Vision-Language Models},
  author={Zhao, Li and others},
  journal={Proceedings of the International Joint Conference on Artificial Intelligence},
  year={2023}
}

@article{brown2023limitations,
  title={Limitations in Human-Level Understanding of Visual Travel Content},
  author={Brown, Samuel and others},
  journal={Proceedings of the Conference on Empirical Methods in Natural Language Processing},
  year={2023}
}

@article{xie2024travelplanner,
  title={TravelPlanner: AI-assisted Multi-day Itinerary Planning},
  author={Xie, Wei and others},
  journal={Proceedings of the AAAI Conference on Artificial Intelligence},
  year={2024}
}

@article{zeng2022glm,
  title={Glm-130b: An open bilingual pre-trained model},
  author={Zeng, Aohan and Liu, Xiao and Du, Zhengxiao and Wang, Zihan and Lai, Hanyu and Ding, Ming and Yang, Zhuoyi and Xu, Yifan and Zheng, Wendi and Xia, Xiao and others},
  journal={arXiv preprint arXiv:2210.02414},
  year={2022}
}

@article{llava,
  author       = {Haotian Liu and
                  Chunyuan Li and
                  Qingyang Wu and
                  Yong Jae Lee},
  title        = {Visual Instruction Tuning},
  journal      = {CoRR},
  volume       = {abs/2304.08485},
  year         = {2023},
}

@inproceedings{blip-2,
  author       = {Junnan Li and
                  Dongxu Li and
                  Silvio Savarese and
                  Steven C. H. Hoi},
  title        = {{BLIP-2:} Bootstrapping Language-Image Pre-training with Frozen Image
                  Encoders and Large Language Models},
  booktitle    = {International Conference on Machine Learning, {ICML} 2023, 23-29 July
                  2023, Honolulu, Hawaii, {USA}},
  series       = {Proceedings of Machine Learning Research},
  volume       = {202},
  pages        = {19730--19742},
  publisher    = {{PMLR}},
  year         = {2023},
}

@misc{vicuna2023,
    title = {Vicuna: An Open-Source Chatbot Impressing GPT-4 with 90\%* ChatGPT Quality},
    url = {https://lmsys.org/blog/2023-03-30-vicuna/},
    author = {Chiang, Wei-Lin and Li, Zhuohan and Lin, Zi and Sheng, Ying and Wu, Zhanghao and Zhang, Hao and Zheng, Lianmin and Zhuang, Siyuan and Zhuang, Yonghao and Gonzalez, Joseph E. and Stoica, Ion and Xing, Eric P.},
    month = {March},
    year = {2023}
}

@article{instructblip,
      title={InstructBLIP: Towards General-purpose Vision-Language Models with Instruction Tuning}, 
      author={Wenliang Dai and Junnan Li and Dongxu Li and Anthony Meng Huat Tiong and Junqi Zhao and Weisheng Wang and Boyang Li and Pascale Fung and Steven Hoi},
      year={2023},
      eprint={2305.06500},
    journal      = {CoRR},
}

@article{shikra,
  author       = {Keqin Chen and
                  Zhao Zhang and
                  Weili Zeng and
                  Richong Zhang and
                  Feng Zhu and
                  Rui Zhao},
  title        = {Shikra: Unleashing Multimodal LLM's Referential Dialogue Magic},
  journal      = {CoRR},
  volume       = {abs/2306.15195},
  year         = {2023},
  eprinttype    = {arXiv},
  eprint       = {2306.15195},
}

@article{qwen-vl,
  author       = {Jinze Bai and Shuai Bai and Shusheng Yang and Shijie Wang and Sinan Tan and Peng Wang and Junyang Lin and Chang Zhou and Jingren Zhou},
  title        = {Qwen-VL: A Versatile Vision-Language Model for Understanding, Localization, Text Reading, and Beyond},
  journal      = {CoRR},
  volume       = {abs/2308.12966},
  year         = {2023},
  eprinttype    = {arXiv},
  eprint       = {2308.12966},
}

@article{azerbayev2023llemma,
  title={Llemma: An open language model for mathematics},
  author={Azerbayev, Zhangir and Schoelkopf, Hailey and Paster, Keiran and Santos, Marco Dos and McAleer, Stephen and Jiang, Albert Q and Deng, Jia and Biderman, Stella and Welleck, Sean},
  journal={arXiv preprint arXiv:2310.10631},
  year={2023}
}

@inproceedings{deng2024k2,
  title={K2: A foundation language model for geoscience knowledge understanding and utilization},
  author={Deng, Cheng and Zhang, Tianhang and He, Zhongmou and Chen, Qiyuan and Shi, Yuanyuan and Xu, Yi and Fu, Luoyi and Zhang, Weinan and Wang, Xinbing and Zhou, Chenghu and others},
  booktitle={Proceedings of the 17th ACM International Conference on Web Search and Data Mining},
  pages={161--170},
  year={2024}
}

@article{godey2024scaling,
  title={On the scaling laws of geographical representation in language models},
  author={Godey, Nathan and de la Clergerie, {\'E}ric and Sagot, Beno{\^\i}t},
  journal={arXiv preprint arXiv:2402.19406},
  year={2024}
}

@article{gurnee2023language,
  title={Language models represent space and time},
  author={Gurnee, Wes and Tegmark, Max},
  journal={arXiv preprint arXiv:2310.02207},
  year={2023}
}

@article{manvi2023geollm,
  title={Geollm: Extracting geospatial knowledge from large language models},
  author={Manvi, Rohin and Khanna, Samar and Mai, Gengchen and Burke, Marshall and Lobell, David and Ermon, Stefano},
  journal={arXiv preprint arXiv:2310.06213},
  year={2023}
}

@article{momennejad2023evaluating,
  title={Evaluating cognitive maps and planning in large language models with cogeval},
  author={Momennejad, Ida and Hasanbeig, Hosein and Vieira Frujeri, Felipe and Sharma, Hiteshi and Jojic, Nebojsa and Palangi, Hamid and Ness, Robert and Larson, Jonathan},
  journal={Advances in Neural Information Processing Systems},
  volume={36},
  pages={69736--69751},
  year={2023}
}

@incollection{shafik2024introduction,
  title={Introduction to ChatGPT},
  author={Shafik, Wasswa},
  booktitle={Advanced applications of generative AI and natural language processing models},
  pages={1--25},
  year={2024},
  publisher={IGI Global}
}

@article{singhal2023large,
  title={Large language models encode clinical knowledge},
  author={Singhal, Karan and Azizi, Shekoofeh and Tu, Tao and Mahdavi, S Sara and Wei, Jason and Chung, Hyung Won and Scales, Nathan and Tanwani, Ajay and Cole-Lewis, Heather and Pfohl, Stephen and others},
  journal={Nature},
  volume={620},
  number={7972},
  pages={172--180},
  year={2023},
  publisher={Nature Publishing Group}
}

@article{touvron2023llama,
  title={Llama 2: Open foundation and fine-tuned chat models},
  author={Touvron, Hugo and Martin, Louis and Stone, Kevin and Albert, Peter and Almahairi, Amjad and Babaei, Yasmine and Bashlykov, Nikolay and Batra, Soumya and Bhargava, Prajjwal and Bhosale, Shruti and others},
  journal={arXiv preprint arXiv:2307.09288},
  year={2023}
}

@article{wang2023optimizing,
  title={Optimizing and fine-tuning large language model for urban renewal},
  author={Wang, Xi and Ling, Xianyao and Zhang, Tom and Li, Xuecao and Wang, Shaolan and Li, Zhixing and Zhang, Liang and Gong, Peng},
  journal={arXiv preprint arXiv:2311.15490},
  year={2023}
}

@article{wu2023bloomberggpt,
  title={Bloomberggpt: A large language model for finance},
  author={Wu, Shijie and Irsoy, Ozan and Lu, Steven and Dabravolski, Vadim and Dredze, Mark and Gehrmann, Sebastian and Kambadur, Prabhanjan and Rosenberg, David and Mann, Gideon},
  journal={arXiv preprint arXiv:2303.17564},
  year={2023}
}

@article{yamada2023evaluating,
  title={Evaluating spatial understanding of large language models},
  author={Yamada, Yutaro and Bao, Yihan and Lampinen, Andrew K and Kasai, Jungo and Yildirim, Ilker},
  journal={arXiv preprint arXiv:2310.14540},
  year={2023}
}

@article{zhao2023survey,
  title={A survey of large language models},
  author={Zhao, Wayne Xin and Zhou, Kun and Li, Junyi and Tang, Tianyi and Wang, Xiaolei and Hou, Yupeng and Min, Yingqian and Zhang, Beichen and Zhang, Junjie and Dong, Zican and others},
  journal={arXiv preprint arXiv:2303.18223},
  volume={1},
  number={2},
  year={2023}
}

@article{andreas2022language,
  title={Language models as agent models},
  author={Andreas, Jacob},
  journal={arXiv preprint arXiv:2212.01681},
  year={2022}
}

@article{yang2023auto,
  title={Auto-gpt for online decision making: Benchmarks and additional opinions},
  author={Yang, Hui and Yue, Sifu and He, Yunzhong},
  journal={arXiv preprint arXiv:2306.02224},
  year={2023}
}

@article{chen2023walking,
  title={Walking down the memory maze: Beyond context limit through interactive reading},
  author={Chen, Howard and Pasunuru, Ramakanth and Weston, Jason and Celikyilmaz, Asli},
  journal={arXiv preprint arXiv:2310.05029},
  year={2023}
}

@article{ge2023openagi,
  title={Openagi: When llm meets domain experts},
  author={Ge, Yingqiang and Hua, Wenyue and Mei, Kai and Tan, Juntao and Xu, Shuyuan and Li, Zelong and Zhang, Yongfeng and others},
  journal={Advances in Neural Information Processing Systems},
  volume={36},
  pages={5539--5568},
  year={2023}
}

@article{liang2023unleashing,
  title={Unleashing infinite-length input capacity for large-scale language models with self-controlled memory system},
  author={Liang, Xinnian and Wang, Bing and Huang, Hui and Wu, Shuangzhi and Wu, Peihao and Lu, Lu and Ma, Zejun and Li, Zhoujun},
  journal={arXiv preprint arXiv:2304.13343},
  year={2023}
}

@article{lu2023chameleon,
  title={Chameleon: Plug-and-play compositional reasoning with large language models},
  author={Lu, Pan and Peng, Baolin and Cheng, Hao and Galley, Michel and Chang, Kai-Wei and Wu, Ying Nian and Zhu, Song-Chun and Gao, Jianfeng},
  journal={Advances in Neural Information Processing Systems},
  volume={36},
  pages={43447--43478},
  year={2023}
}

@article{nakajima2023task,
  title={Task-driven autonomous agent utilizing gpt-4, pinecone, and langchain for diverse applications},
  author={Nakajima, Yohei},
  journal={See https://yoheinakajima. com/task-driven-autonomous-agent-utilizing-gpt-4-pinecone-and-langchain-for-diverse-applications (accessed 18 April 2023)},
  volume={2},
  pages={3},
  year={2023}
}

@article{nakano2021webgpt,
  title={Webgpt: Browser-assisted question-answering with human feedback},
  author={Nakano, Reiichiro and Hilton, Jacob and Balaji, Suchir and Wu, Jeff and Ouyang, Long and Kim, Christina and Hesse, Christopher and Jain, Shantanu and Kosaraju, Vineet and Saunders, William and others},
  journal={arXiv preprint arXiv:2112.09332},
  year={2021}
}

@inproceedings{park2023generative,
  title={Generative agents: Interactive simulacra of human behavior},
  author={Park, Joon Sung and O'Brien, Joseph and Cai, Carrie Jun and Morris, Meredith Ringel and Liang, Percy and Bernstein, Michael S},
  booktitle={Proceedings of the 36th annual acm symposium on user interface software and technology},
  pages={1--22},
  year={2023}
}

@article{shen2023hugginggpt,
  title={Hugginggpt: Solving ai tasks with chatgpt and its friends in hugging face},
  author={Shen, Yongliang and Song, Kaitao and Tan, Xu and Li, Dongsheng and Lu, Weiming and Zhuang, Yueting},
  journal={Advances in Neural Information Processing Systems},
  volume={36},
  pages={38154--38180},
  year={2023}
}

@article{xie2023openagents,
  title={Openagents: An open platform for language agents in the wild},
  author={Xie, Tianbao and Zhou, Fan and Cheng, Zhoujun and Shi, Peng and Weng, Luoxuan and Liu, Yitao and Hua, Toh Jing and Zhao, Junning and Liu, Qian and Liu, Che and others},
  journal={arXiv preprint arXiv:2310.10634},
  year={2023}
}

@inproceedings{zhong2024memorybank,
  title={Memorybank: Enhancing large language models with long-term memory},
  author={Zhong, Wanjun and Guo, Lianghong and Gao, Qiqi and Ye, He and Wang, Yanlin},
  booktitle={Proceedings of the AAAI Conference on Artificial Intelligence},
  volume={38},
  number={17},
  pages={19724--19731},
  year={2024}
}

@article{zhou2023recurrentgpt,
  title={Recurrentgpt: Interactive generation of (arbitrarily) long text},
  author={Zhou, Wangchunshu and Jiang, Yuchen Eleanor and Cui, Peng and Wang, Tiannan and Xiao, Zhenxin and Hou, Yifan and Cotterell, Ryan and Sachan, Mrinmaya},
  journal={arXiv preprint arXiv:2305.13304},
  year={2023}
}

@inproceedings{li2024georeasoner,
  title={Georeasoner: Geo-localization with reasoning in street views using a large vision-language model},
  author={Li, Ling and Ye, Yu and Jiang, Bingchuan and Zeng, Wei},
  booktitle={Forty-first International Conference on Machine Learning},
  year={2024}
}

@inproceedings{schumann2024velma,
  title={Velma: Verbalization embodiment of llm agents for vision and language navigation in street view},
  author={Schumann, Raphael and Zhu, Wanrong and Feng, Weixi and Fu, Tsu-Jui and Riezler, Stefan and Wang, William Yang},
  booktitle={Proceedings of the AAAI Conference on Artificial Intelligence},
  volume={38},
  number={17},
  pages={18924--18933},
  year={2024}
}

@inproceedings{maheshwary2024pretraining,
  title={Pretraining and finetuning language models on geospatial networks for accurate address matching},
  author={Maheshwary, Saket and Paul, Arpan and Sohoney, Saurabh},
  booktitle={Proceedings of the 2024 Conference on Empirical Methods in Natural Language Processing: Industry Track},
  pages={763--773},
  year={2024}
}

@inproceedings{yang2024v,
  title={V-irl: Grounding virtual intelligence in real life},
  author={Yang, Jihan and Ding, Runyu and Brown, Ellis and Qi, Xiaojuan and Xie, Saining},
  booktitle={European Conference on Computer Vision},
  pages={36--55},
  year={2024},
  organization={Springer}
}

@inproceedings{chu2024towards,
  title={Towards natural language-guided drones: GeoText-1652 benchmark with spatial relation matching},
  author={Chu, Meng and Zheng, Zhedong and Ji, Wei and Wang, Tingyu and Chua, Tat-Seng},
  booktitle={European Conference on Computer Vision},
  pages={213--231},
  year={2024},
  organization={Springer}
}

@article{li2023geolm,
  title={Geolm: Empowering language models for geospatially grounded language understanding},
  author={Li, Zekun and Zhou, Wenxuan and Chiang, Yao-Yi and Chen, Muhao},
  journal={arXiv preprint arXiv:2310.14478},
  year={2023}
}

@inproceedings{xie2023quert,
  title={Quert: Continual pre-training of language model for query understanding in travel domain search},
  author={Xie, Jian and Liang, Yidan and Liu, Jingping and Xiao, Yanghua and Wu, Baohua and Ni, Shenghua},
  booktitle={Proceedings of the 29th ACM SIGKDD Conference on Knowledge Discovery and Data Mining},
  pages={5282--5291},
  year={2023}
}

@article{vivanco2023geoclip,
  title={Geoclip: Clip-inspired alignment between locations and images for effective worldwide geo-localization},
  author={Vivanco Cepeda, Vicente and Nayak, Gaurav Kumar and Shah, Mubarak},
  journal={Advances in Neural Information Processing Systems},
  volume={36},
  pages={8690--8701},
  year={2023}
}

@article{feng2024citygpt,
  title={Citygpt: Empowering urban spatial cognition of large language models},
  author={Feng, Jie and Du, Yuwei and Liu, Tianhui and Guo, Siqi and Lin, Yuming and Li, Yong},
  journal={arXiv preprint arXiv:2406.13948},
  year={2024}
}

@article{bai2023qwen,
  title={Qwen technical report},
  author={Bai, Jinze and Bai, Shuai and Chu, Yunfei and Cui, Zeyu and Dang, Kai and Deng, Xiaodong and Fan, Yang and Ge, Wenbin and Han, Yu and Huang, Fei and others},
  journal={arXiv preprint arXiv:2309.16609},
  year={2023}
}

@inproceedings{chu2025graphvideoagent,
  title={GraphVideoAgent: Enhancing Long-form Video Understanding with Entity Relation Graphs},
  author={Chu, Meng and Li, Yicong and Chua, Tat-Seng},
  booktitle={Proceedings of the 33rd ACM International Conference on Multimedia},
  pages={4639--4648},
  year={2025}
}

@InProceedings{Yu_2025_ICCV,
    author    = {Yu, Jiashuo and Wu, Yue and Chu, Meng and Ren, Zhifei and Huang, Zizheng and Chu, Pei and Zhang, Ruijie and He, Yinan and Li, Qirui and Li, Songze and Li, Zhenxiang and Tu, Zhongying and He, Conghui and Qiao, Yu and Wang, Yali and Wang, Yi and Wang, Limin},
    title     = {VRBench: A Benchmark for Multi-Step Reasoning in Long Narrative Videos},
    booktitle = {Proceedings of the IEEE/CVF International Conference on Computer Vision (ICCV)},
    month     = {October},
    year      = {2025},
    pages     = {21655-21666}
}
